\documentclass[aip,pop,reprint,amsmath,amssymb]{revtex4-2}
\usepackage{graphicx}
\usepackage{booktabs}
\usepackage{xcolor}
\usepackage[colorlinks=true,linkcolor=blue,citecolor=blue,urlcolor=blue]{hyperref}

\begin{document}

\title{A trust layer for fast pedestal-stability surrogates: calibrated error bars,
out-of-distribution alarms and few-shot recalibration}

\author{Amin Vakhshouri}
\email{amin@dataquanta.io}
\affiliation{dataquanta LLC, Delaware, USA}

\date{\today}

\begin{abstract}
Machine-learning surrogates of linear peeling-ballooning stability now run in
milliseconds, but they are mostly deployed as point estimators: they do not say how
wrong they may be, or when a query lies outside what they were trained on. For a fusion
company designing a new device, every equilibrium of interest is outside the training data
of existing surrogates, so this is the case that matters. We describe a model-agnostic trust
layer for stability surrogates, made of conformal error bars, out-of-distribution alarms that
route doubtful cases to the physics code, and recalibration from a few simulations of a new
design, and we test it with pre-registered hypotheses. On a large public dataset of JET-like
equilibria with ideal-MHD growth rates, the error bars hold their nominal coverage in
distribution and fall short under a shift in plasma shape; a few dozen simulations of the new
shape, or routing the cases the alarms flag, restore it. On resistive BOUT++ surrogates, the
bars stay calibrated on held-out runs, the alarm flags tokamaks absent from training, and a
scenario scan is screened in seconds instead of hours. Across equilibria from the SPARC library
and the DIII-D coil set, a surrogate trained on our own graded BOUT++ labels beats the baseline
on unseen equilibria, while a surrogate trained elsewhere ranks the same cases backwards and its
alarm declines them. Building our own neural-operator surrogate, the harness exposed two input
defects whose fix closed its gap to a simple network. For every surrogate tested, the trust layer
keeps coverage near target, with a bar width set by that surrogate's own error. Not every
pre-registered hypothesis held, and we report those results and the limits of all of them.
\end{abstract}

\maketitle

\section{Introduction}

Linear stability of the edge pedestal to peeling-ballooning modes sets the pedestal height in
predictive models such as EPED~\cite{snyder2009eped} and constrains the operating space of
present and planned tokamaks. Codes that compute it (ELITE, MISHKA, BOUT++, PB3D) need
minutes to hours per equilibrium~\cite{dudson2009bout,weyens2017pb3d}, so scenario scans and
between-shot analysis are rationed by compute. Machine-learning surrogates remove that
bottleneck: EPED1-NN~\cite{meneghini2017epednn}, EuroPED-NN~\cite{panera2024europednn},
MISHKA-NN~\cite{heo2024mishkann}, KARHU~\cite{bruncrona2025karhu} and, most recently,
ELMO~\cite{li2026elmo} predict pedestal or stability quantities in milliseconds.

Speed alone does not make a surrogate usable for decisions. A physicist or an engineer
will act on a growth rate only if it comes with an error bar that means what it says, and
with a warning when the input is unlike the training data. Existing pedestal surrogates
either return point estimates (KARHU, MISHKA-NN, EPED1-NN) or Bayesian uncertainties whose
calibration is not guaranteed: ELMO's authors report that their Gaussian-process intervals
cover 62\% of test cases at a nominal 68\% and 86\% at a nominal 95\%~\cite{li2026elmo}.
The problem is sharpest for fusion companies designing new devices, whose equilibria are by
construction outside the training distribution of every existing surrogate.

This paper treats the missing piece, a \emph{trust layer}, as the object of study. The layer
wraps any surrogate and supplies (i) distribution-free conformal intervals with a finite-sample
coverage guarantee under exchangeability~\cite{vovk2005algorithmic,lei2018distribution},
(ii) out-of-distribution (OOD) alarms that route unfamiliar cases to the physics code, and
(iii) cheap recalibration on a handful of labelled cases from a new device. Conformal
prediction has been applied to fusion field surrogates by Gopakumar
\emph{et~al.}~\cite{gopakumar2026conformal,gopakumar2025physics}; to our knowledge this is its
first systematic test on a stability surrogate, together with shift, alarms and recalibration.

Contributions:
\begin{enumerate}
\item A pre-registered evaluation of the trust layer on the public KARHU dataset
(38,879 equilibria)~\cite{karhu2026data}, under a controlled plasma-shape shift, with all
outcomes reported.
\item Two practical onboarding paths for a new device: recalibration on $\sim$30 labelled cases,
or zero-label routing of flagged cases to the physics code.
\item A resistive BOUT++ case study with real-machine equilibria: calibrated intervals,
an ensemble alarm that separates seen from unseen tokamaks, and a decision-level scenario scan.
\item The same harness applied to our own architecture, which exposed two input defects (a
radial grid too coarse at the pedestal and unscaled profiles) and confirmed their fix on a
pre-registered shifted test.
\item A pre-registered onboarding test (how many simulations of a new design, chosen how) and
a path from a G-EQDSK file to a calibrated answer or a reasoned refusal.
\item An account of what went wrong along the way (Sec.~\ref{sec:lessons}), including a
reported number we could not reproduce.
\end{enumerate}

\section{Trust layer}

Let $\hat\gamma(x)$ be any surrogate of the growth rate $\gamma$ for input $x$.

\paragraph*{Split conformal intervals.} Given $n$ labelled calibration cases not used for
training, with scores $s_i = |\gamma_i - \hat\gamma(x_i)|$, the interval
$\hat\gamma(x) \pm q$ with $q$ the $\lceil (n+1)(1-\alpha)\rceil$-th smallest score covers a new
exchangeable case with probability at least $1-\alpha$~\cite{lei2018distribution}. We use
$\alpha = 0.1$. The guarantee is marginal, not conditional on $x$.

\paragraph*{Mondrian (per-range) calibration.} Computing $q$ separately within bins of the
\emph{predicted} growth rate gives coverage per bin~\cite{vovk2005algorithmic}, which matters
because errors are not uniform across stable, marginal and strongly unstable cases.

\paragraph*{Weighted conformal under covariate shift.} When the test distribution differs from
calibration by a known or estimable likelihood ratio $w(x)$, weighting the calibration scores
restores coverage~\cite{tibshirani2019covariate}. We estimate $w$ with a logistic domain
classifier on shape features. Weighted conformal requires overlapping support: where the
calibration data contain no comparable cases it returns infinite intervals, which is the
correct answer.

\paragraph*{Few-shot recalibration.} With $m$ labelled cases from a new device, split conformal
on those $m$ cases alone gives a valid interval for $m \ge 9$ at $\alpha=0.1$; the open
question is how its width and its draw-to-draw variability depend on $m$.

\paragraph*{OOD alarms and routing.} We use two alarms: the spread $\sigma(x)$ of a $K=5$ deep
ensemble~\cite{lakshminarayanan2017ensembles}, and a Mahalanobis distance of the input from the
training data~\cite{lee2018mahalanobis}. Each threshold is set from in-distribution calibration
data only (99th percentile, or the maximum when $n$ is small), so the false-alarm rate is fixed
before any shifted data are seen. Flagged cases are sent to the physics code; the coverage that
matters is then the coverage of the cases kept.

\section{Data and surrogates}

\paragraph*{KARHU (ideal MHD).} The public KARHU training set~\cite{karhu2026data} contains
38,880 JET-ILW-like equilibria with MISHKA growth rates for toroidal mode numbers 3--50,
pressure, safety-factor and $RB_\phi$ profiles on 64 flux points, a 128-point boundary, and
global scalars. The target is the maximum growth rate $\gamma_{\max} \in [0, 0.59]$ (27.4\%
stable); one case without a label is dropped. The boundary is normalised and up-down symmetric;
elongation barely varies (interquartile range within 1.67), triangularity $\delta$ spans
0.02--0.59, so $\delta$ is the shift variable.

\paragraph*{BOUT++ (resistive MHD).} Linear growth rates of the BOUT++ \texttt{elm\_pb}
model~\cite{dudson2009bout} on the CBM18 equilibrium, varying resistivity $\eta$ and viscosity
$\nu$: 50 training runs on an $\eta\times\nu$ grid ($\eta \in [10^{-6}, 2\times10^{-4}]$,
$\nu \in [3\times10^{-7}, 3\times10^{-4}]$) and 42 held-out runs at geometric midpoints of that grid,
generated afterwards. Four real-machine equilibria (three SPARC, one DIII-D) are used in
fine-tuning; JET 79692, NSTX 132588 and ITER 15\,MA are never seen.

\paragraph*{Surrogates.} \emph{MLP:} a 4-layer multilayer perceptron on the 333 KARHU inputs,
used to test the trust layer independently of architecture. \emph{plasmafast:} a
flux-surface attention network (0.80\,M-parameter encoder) that maps an
equilibrium and $(\eta, \nu)$ to $\gamma$. Its input is built from the $q(\psi)$ and $p(\psi)$
profiles on a $64\times64$ flux grid, with pressure rescaled to a fixed axis value; it does not
see plasma shape or the pressure magnitude. For BOUT++ we feed $\log_{10}\eta$ and
$\log_{10}\nu$ directly to the growth-rate head (Sec.~\ref{sec:lessons}). For KARHU (``v2'') we
feed 15 global quantities ($\beta_N$, $\beta_p$, $q_{95}$, $q_0$, $\kappa$, $\delta$, inverse
aspect ratio, field, current, axis pressure and others) to the head and train from scratch.

\section{Results on KARHU (pre-registered)}\label{sec:karhu}

\begin{figure*}
\includegraphics[width=\textwidth]{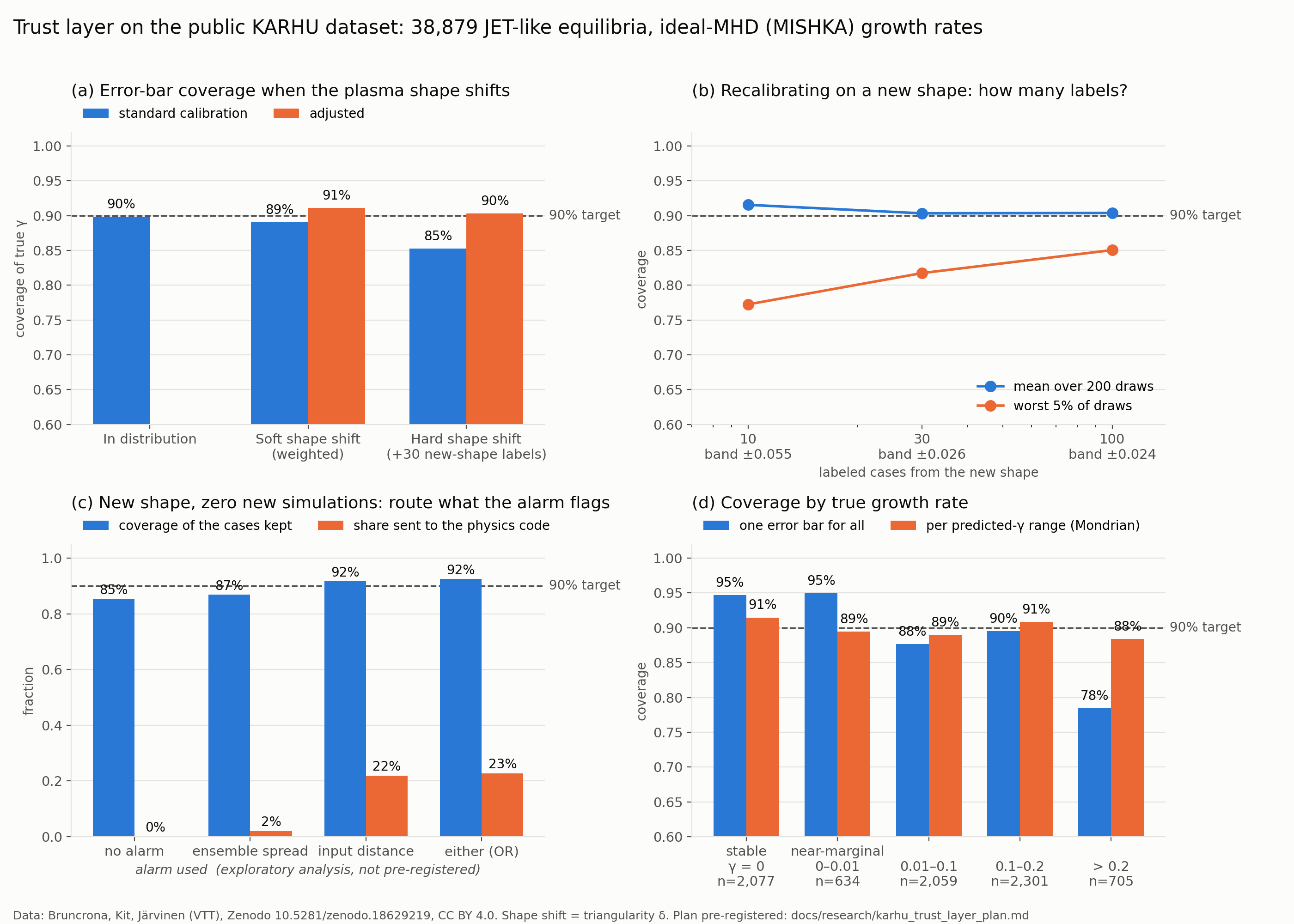}
\caption{Trust layer on KARHU (MLP ensemble, $K=5$). (a) Coverage in distribution and under
a soft and a hard triangularity shift, with standard and adjusted calibration. (b) Few-shot
recalibration on the new shape: mean and worst-5\% coverage over 200 draws, and interval
half-width. (c) Exploratory: coverage of the cases kept after routing what each alarm flags.
(d) Coverage by true growth rate, one interval for all vs.\ per predicted-$\gamma$ range.}
\label{fig:karhu}
\end{figure*}

The hypotheses, pass criteria and one design amendment were committed before any result
(repository file \texttt{docs/research/karhu\_trust\_layer\_plan.md}). Splits: S0 random
60/20/20; S1 hard shift, training and calibration on $\delta \le 0.356$ (70th percentile),
test on the rest ($n = 11{,}664$); S2 soft shift, a test set drawn with probability
$\propto e^{2 z_\delta}$, overlapping the source. Figure~\ref{fig:karhu} summarises.

\begin{table}
\caption{Pre-registered outcomes on KARHU (MLP ensemble).}
\label{tab:karhu}
\begin{ruledtabular}
\begin{tabular}{lp{0.58\columnwidth}l}
 & Result & Verdict \\
\hline
H1 & 89.9\% at 90\%, $q = \pm0.016$, $R^2 = 0.933$ & pass \\
H2 & hard shift: 85.3\% (criterion $<85\%$) & not met \\
H3 & soft shift: 89.0\% $\to$ 91.1\% weighted & pass (weak) \\
H4 & $m = 10/30/100$: 91.5/90.3/90.4\%, $\pm0.055/0.026/0.024$ & knee at 30 \\
H5 & AUROC: spread 0.63, distance 0.73 & mixed \\
H6 & $\gamma > 0.2$: 78.4\% $\to$ 88.4\% (Mondrian) & pass \\
\end{tabular}
\end{ruledtabular}
\end{table}

\paragraph*{In distribution (H1).} Coverage is 89.9\% at a 90\% target with a half-width of 0.016.

\paragraph*{Shape shift (H2, H3).} Training on lower triangularity and testing on higher lowers
coverage from 90.1\% on held-out source cases to 85.3\%, and raises the mean absolute error from 0.0083 to 0.0106.
We had predicted a larger drop; within this JET-like family the shift is mild. On the soft shift,
standard calibration already gives 89.0\% and weighting gives 91.1\%, with 7.7\% infinite
intervals; on the hard shift weighting returns infinite intervals for 99.7\% of cases, as
expected without support overlap.

\paragraph*{Onboarding a new shape (H4).} Recalibrating on $m$ labelled cases of the new shape
gives mean coverage 91.5/90.3/90.4\% for $m = 10/30/100$, with half-widths 0.055/0.026/0.024.
Thirty cases bring the interval close to its in-distribution width; one hundred add little.
Coverage varies between draws of the $m$ cases: the worst 5\% of 30-case draws reach 82\%.

\paragraph*{Alarms (H5) and routing (exploratory).} As detectors of the shifted shape, the
ensemble spread is weak (AUROC 0.63) and the Mahalanobis distance moderate (0.73), each at
about 1\% false alarms. In an analysis we had not pre-registered we asked the question a user
needs answered: does the alarm fire on the cases the surrogate gets wrong? The spread ranks
misses well (AUROC 0.86; 92\% of the cases it flags lie outside the interval) but fires on only
2\% of shifted cases; the distance flags 22\%. Routing the union (22.7\% of shifted cases, 1.9\%
false alarms) to the physics code leaves 92.5\% coverage on the cases kept, without any new
simulation.

\paragraph*{Coverage by growth rate (H6).} One interval for all gives 94.7\% for stable cases
and 95.0\% near marginal stability but only 78.4\% for $\gamma > 0.2$. Calibrating per predicted
range gives 88--92\% in every range, with half-widths from $\pm0.007$ near marginal stability to
$\pm0.023$ for strongly unstable cases.

\subsection*{The deployed architecture on KARHU}

\begin{figure*}
\includegraphics[width=\textwidth]{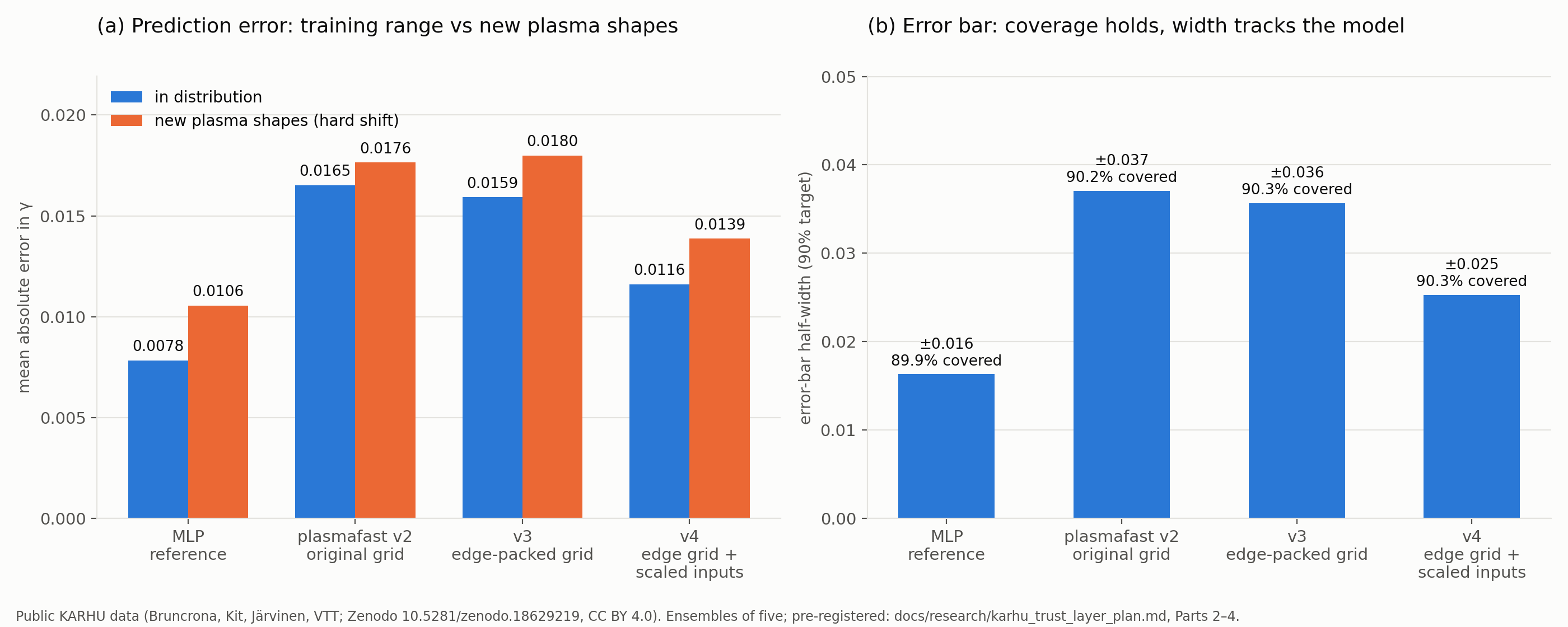}
\caption{\label{fig:models} The deployed architecture through the same trust layer on KARHU.
(a) Mean absolute error in distribution (S0) and on the shifted shapes (S1 target).
(b) Calibrated half-width and in-distribution coverage at a 90\% target.}
\end{figure*}

We then trained plasmafast itself on KARHU and passed it through the same harness, in three
pre-registered stages (Parts 2--4 of the plan) plus diagnostics that we label exploratory.
Table~\ref{tab:models} and Fig.~\ref{fig:models} compare all surrogates.

\paragraph*{v2: global inputs, original grid.} Feeding the 15 global quantities to the
growth-rate head matters: a single model without them reaches $R^2 = 0.740$, with them 0.875
(pre-registered V2, passed). The five-member ensemble reaches $R^2 = 0.880$ in distribution,
short of the MLP's 0.933 (V1, failed). The trust layer behaves as on the MLP (coverage 90.2\%,
weighted coverage under the soft shift 90.4\%, 30-case recalibration 90.6\%), with intervals
about twice as wide because the errors are.

\paragraph*{Why plasmafast trailed (exploratory).} An MLP given exactly plasmafast's
information (the $q$ and $p$ profiles and the 15 scalars) reaches $R^2 = 0.92$--0.94, so the
information is present. Two losses explain the gap. First, the grid: resampling the profiles
onto plasmafast's 64-point uniform-$\rho$ grid raises the MLP's error by 16\% (three seeds),
because the pedestal falls between grid points (Sec.~\ref{sec:lessons}); float16 storage costs
nothing. Second, scaling: plasmafast reads the $q$ and $p$ channels raw, while the MLP
standardizes every input. In a two-by-two test of single networks (same seed), standardizing
$q$ and $p$ per radial point lowers the error by 18\% on the original grid ($R^2$ 0.875 to
0.900); the edge-packed grid below adds nothing on raw inputs but a further 11\% once inputs
are standardized ($R^2$ 0.924).

\paragraph*{v3: edge-packed grid.} Regenerating the frames on a 96-point grid with 64 points in
$\psi_N \in [0.8, 1]$ (edge spacing 0.0032) alone changes little for the network: $R^2 = 0.884$,
mean error 4\% below v2, against a pre-registered 10\% (W1 and W2 failed).

\paragraph*{v4: edge-packed grid and standardized profiles.} Because standardization was chosen
after seeing one model's in-distribution test error, we pre-registered the shifted split S1,
which played no part in that choice, as the confirmatory test. On S1's new shapes the v4
ensemble's mean error is 23\% below v3's (0.0139 against 0.0180; criterion 10\%, passed). In
distribution it reaches $R^2 = 0.9325$, level with the MLP (0.933) but 0.0005 short of the bar
we had set (failed); its mean error, 0.0116, is 30\% below v2's yet still above the MLP's
0.0078. Its intervals narrow accordingly ($\pm0.025$ against $\pm0.037$ for v2), and
calibration per predicted range covers strongly unstable cases at 88.4\%, as for the MLP.

\paragraph*{What carries over.} For every surrogate, from 0.880 to 0.933 in $R^2$, the conformal
layer keeps coverage at 90--91\% in distribution and restores it under shift; what changes is
the interval width, which tracks the surrogate's own error. A better surrogate buys a tighter
interval; a worse one never buys a false guarantee.

\begin{table}
\caption{\label{tab:models} Same trust-layer harness, four surrogates on KARHU (ensembles of five;
S0 in distribution, S1 hard and S2 soft triangularity shift; p5: worst 5\% of 200
draws of the 30 recalibration cases).}
\begin{ruledtabular}
\footnotesize\setlength{\tabcolsep}{2.5pt}
\begin{tabular}{lcccc}
 & MLP & v2 & v3 & v4 \\
\hline
$R^2$ (S0) & 0.9333 & 0.8802 & 0.8836 & 0.9325 \\
MAE (S0) & 0.0078 & 0.0165 & 0.0159 & 0.0116 \\
Coverage (S0) & 89.9\% & 90.2\% & 90.3\% & 90.3\% \\
Half-width (S0) & 0.016 & 0.037 & 0.036 & 0.025 \\
MAE, S1 target & 0.0106 & 0.0176 & 0.0180 & 0.0139 \\
Coverage, S1 target & 85.3\% & 88.0\% & 87.0\% & 86.1\% \\
Weighted coverage, S2 & 91.1\% & 90.4\% & 91.3\% & 90.7\% \\
30-case recal. (p5) & 90.3 (82) & 90.6 (81) & 90.4 (81) & 90.6 (79) \\
$\gamma>0.2$, per range & 88.4\% & 83.1\% & 84.1\% & 88.4\% \\
\end{tabular}
\end{ruledtabular}
\end{table}

\subsection*{Onboarding a new design}\label{sec:onboarding}

The customer's question is how many simulations of a new design a surrogate needs, and whether
letting the trust layer choose which ones to run saves any. We pre-registered a test (Part~5 of
the plan): the v4 ensemble trained on the old family (S1 source) meets the high-triangularity
family; a fixed 40\% of that family is the test set and the other 6,999 equilibria form a pool
whose labels are revealed on request. In every run 30 pool cases drawn at random are used only to
calibrate the bar, because cases chosen by the model are not exchangeable with the test set and
cannot calibrate it. The rest of the budget is chosen at random or guided by the ensemble (highest
spread, then spread out in parameter space), and the members are fine-tuned after each round.

\begin{figure*}
\includegraphics[width=\textwidth]{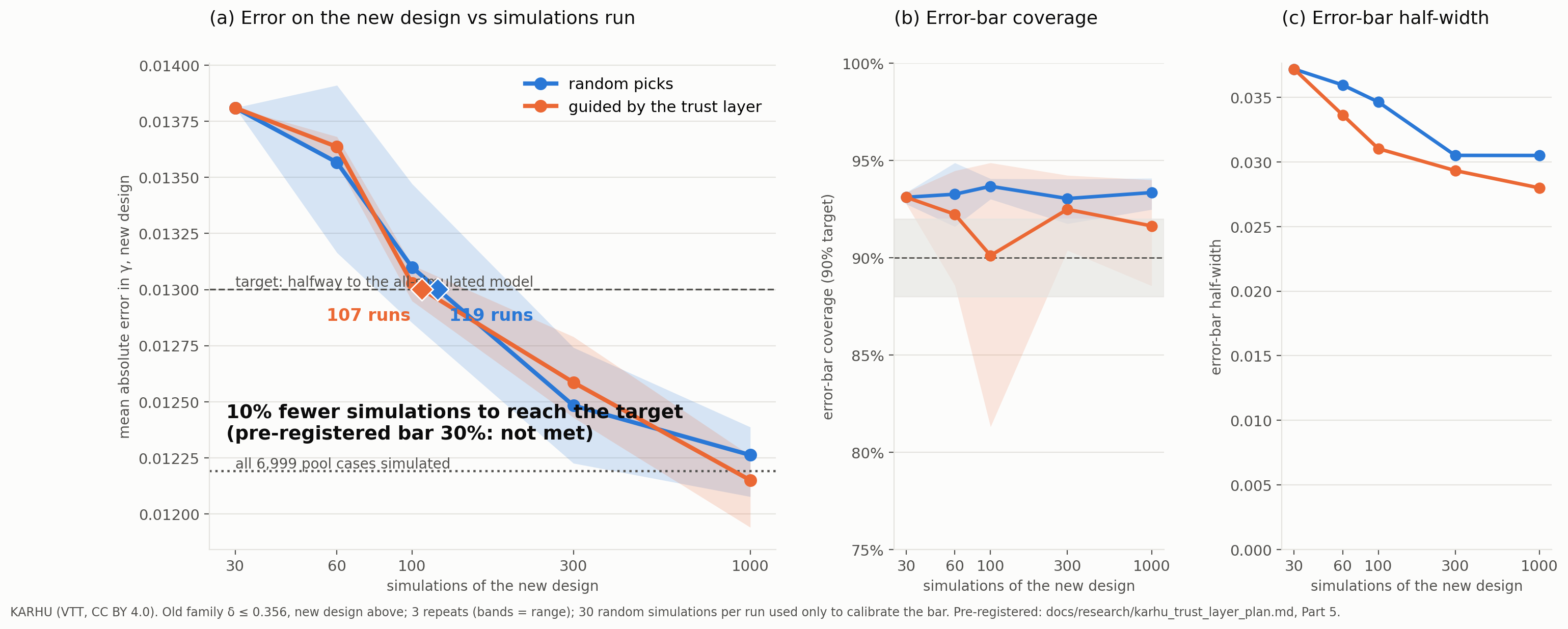}
\caption{\label{fig:onboarding} Onboarding the high-triangularity family (pre-registered; three
repeats, bands show the range). (a) Error on the new family's test set against the number of new
simulations, random against guided selection; diamonds mark where each reaches the target halfway
between the base model and the model trained on the whole pool. (b) Coverage of the conformal bar
calibrated on 30 random simulations. (c) Its half-width.}
\end{figure*}

Within this family the model transfers well: with the 30 calibration runs alone its error on the
new shapes is 0.0138, against 0.0122 when all 6,999 pool cases are simulated (Fig.~\ref{fig:onboarding}).
To close half of that gap, guided selection needs 107 simulations and random 119, a 10\% saving
against the 30\% we had set as the bar (not met). Guided selection is never worse than random
(passed), and the bar stays honest throughout, on the conservative side: mean coverage is 93--94\% for
random and 90--93\% for guided picks, above the 88--92\% band we had set (not met), with one guided
repeat falling to 81\%. In an analysis we had not pre-registered, 1,000 guided simulations match
the model trained on all 6,999 (mean error 0.0122 for both; $R^2$ 0.876 against 0.864), and the
guided picks cover the sparse corners of the new family that random picks miss. For a mild shift
the practical answer is therefore about 30 simulations for an honest bar; guided selection pays
when the budget is larger.

\subsection*{From an equilibrium file to an answer}\label{sec:predict}

A tool must accept the file a user has. We built one path from a G-EQDSK to plasmafast's inputs:
the boundary is taken at $\psi_N = 0.99$ (KARHU equilibria have a fixed boundary), $q$ and the
normalized pressure are sampled on KARHU's flux grid, and the 15 global inputs are computed by their
standard definitions, checked against KARHU's own data where it allows (edge $q$ exactly;
normalized current and axis field within 3\% and 2\%). The frames come from the same function
that built the training frames, which regenerates the training set byte for byte, so training and
inference cannot drift apart (Sec.~\ref{sec:lessons}). The deployable ensemble (v4 with a floor on
the per-point spread used for standardization, Sec.~\ref{sec:lessons}) reaches $R^2 = 0.933$ and
90\% coverage at $\pm0.026$ on KARHU, and reproduces that accuracy through this path.

\begin{figure*}
\includegraphics[width=\textwidth]{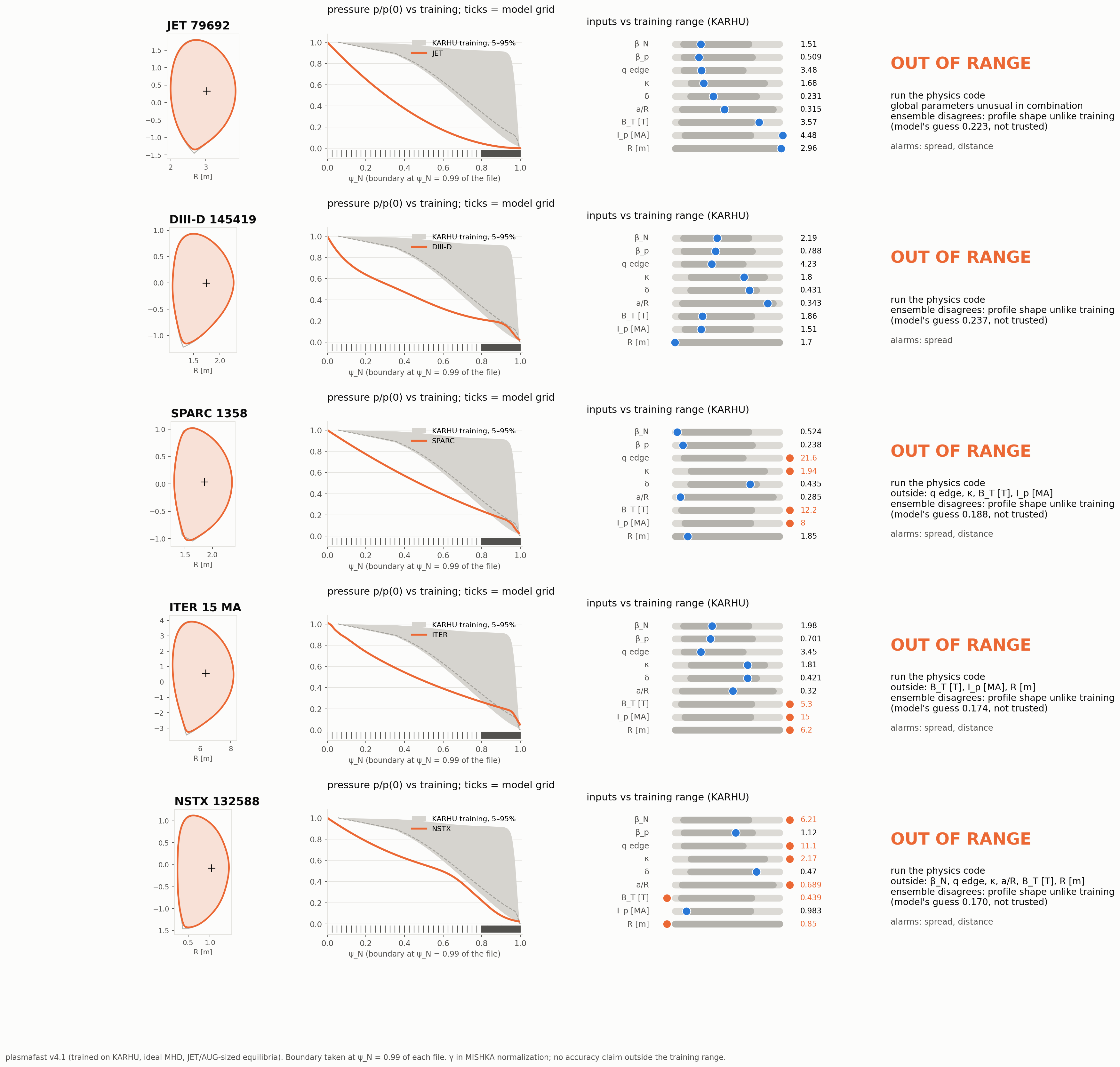}
\caption{\label{fig:cards} The deployable surrogate on five real G-EQDSK equilibria. Columns:
plasma boundary used; pressure profile against the KARHU training envelope (5--95\%), with the
model's radial grid as ticks; global inputs against their training range (dark: 5--95\%, light:
0.5--99.5\%); verdict and reasons.}
\end{figure*}

We then asked it about five real equilibria (Fig.~\ref{fig:cards}). It answers for none of them.
SPARC, ITER and NSTX lie outside the training range in field, current, size or shape, which a
range check on the global inputs already shows. JET and DIII-D are subtler: DIII-D's global inputs
all lie inside the training range, yet the five members disagree (spread 0.053 against a threshold
of 0.030). The reason is visible in the profiles: every real pressure profile is more peaked than
KARHU's training set from mid-radius outward ($p/p_0 = 0.32$ at $\psi_N = 0.6$ for DIII-D and
0.17 for JET, against 0.59 for KARHU's 5th percentile). The ensemble caught a shift that a check on scalar
inputs would have missed, and the tool reports it as a reason, not just a flag. A surrogate that
is to answer for these machines needs training equilibria with realistic profile shapes.

\section{Results on BOUT++}

\paragraph*{Calibrated intervals.} On the 42 held-out runs, the surrogate's mean absolute
error is 0.028 and split conformal gives 92.3\% coverage at a 90\% target with a half-width of
0.052--0.054. On the CBM18 reference case ($\eta = 10^{-4}$, $\nu = 3\times10^{-7}$, BOUT++
$\gamma = 0.423$) the prediction is 0.413 ($-2.5\%$). As a code-to-code check of BOUT++ itself,
the near-ideal run ($\eta = 10^{-6}$) gives 0.294 against PB3D's ideal-MHD 0.245 on the same
equilibrium (+20\%).

\paragraph*{Alarm on real machines.} Table~\ref{tab:ood} lists the spread of a five-member head
ensemble for six real equilibria, relative to the largest spread on the in-distribution
held-out set. All six inputs are built with the inference pipeline (Sec.~\ref{sec:lessons}). All three unseen tokamaks are flagged, including JET, where the surrogate is
badly wrong (0.11 vs.\ BOUT++ 1.81); none of the three fine-tuning equilibria are. With six
cases this is a demonstration, not a rate. An input-distance alarm, useful on KARHU, flags all
six here, because the training data contain essentially one geometry.

\begin{table}
\caption{Ensemble-spread alarm on real equilibria (BOUT++ surrogate). Threshold: largest spread
on 42 in-distribution held-out runs.}
\label{tab:ood}
\begin{ruledtabular}
\begin{tabular}{lccc}
Equilibrium & In training & Spread / mean & Alarm \\
\hline
SPARC 1358 & yes & 1.0$\times$ & -- \\
SPARC 1514 & yes & 1.1$\times$ & -- \\
DIII-D 145419 & yes & 0.6$\times$ & -- \\
JET 79692 & no & 5.9$\times$ & yes \\
NSTX 132588 & no & 9.9$\times$ & yes \\
ITER 15\,MA & no & 4.8$\times$ & yes \\
\end{tabular}
\end{ruledtabular}
\end{table}

\begin{figure*}
\includegraphics[width=\textwidth]{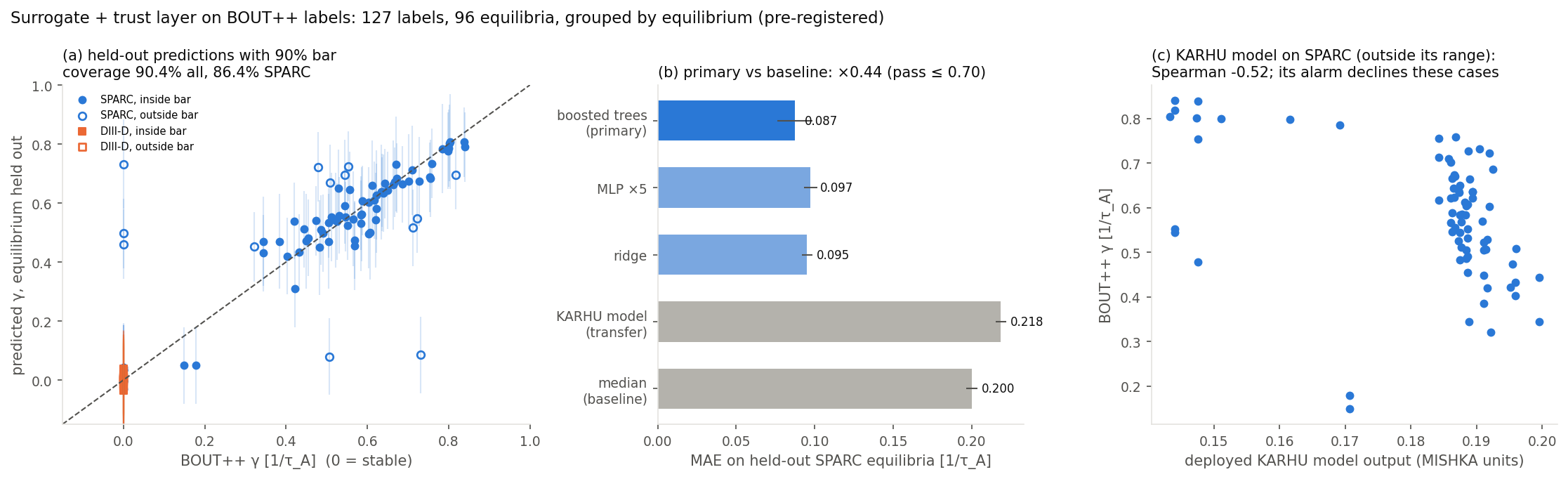}
\caption{Surrogate and trust layer on BOUT++ labels, equilibria held out (pre-registered). Left:
held-out predictions with the 90\% CV+ interval. Centre: error on held-out SPARC equilibria by
model. Right: the KARHU-trained surrogate's output on the same SPARC cases, outside its range.}
\label{fig:boutlabels}
\end{figure*}

\paragraph*{Many equilibria.} The results above rest on one equilibrium family (CBM18 and four
fine-tuning cases). To test the recipe across equilibria, we graded BOUT++ elm\_pb\_tor growth rates
on 96 equilibria from the public SPARC library and from FreeGS solutions on the DIII-D coil
set (127 runs), keeping only runs whose seeded mode is resolved (radial width $\geq 5$ cells),
fitted with $R^2 \geq 0.8$, and not dominated by a numerical mode at the radial boundary
(Sec.~\ref{sec:lessons}). In a pre-registered test with whole equilibria held out (grouped
five-fold cross-validation, five fold assignments), gradient-boosted trees on the 15 global inputs,
$\log_{10}\eta$ and three pedestal metrics reach a mean absolute error of 0.087 on SPARC
equilibria against 0.200 for the median (ratio 0.44; pre-registered pass $\leq 0.70$).
Group-wise CV+ 90\% intervals cover 90.4\% of all held-out labels and
86.4\% of SPARC labels, with a median half-width of
0.20 (Fig.~\ref{fig:boutlabels}). The KARHU-trained surrogate
ranks the same SPARC growth rates backwards (Spearman $-0.52$): SPARC lies
outside KARHU's range, and its alarm declines these cases (Sec.~\ref{sec:predict}). All DIII-D labels so
far are stable, so the signal is SPARC's; the stable cases have pedestal pressure gradients
$\alpha_{\max} \lesssim 8$, the unstable SPARC cases $\alpha_{\max} \geq 15$.

\begin{figure*}
\includegraphics[width=\textwidth]{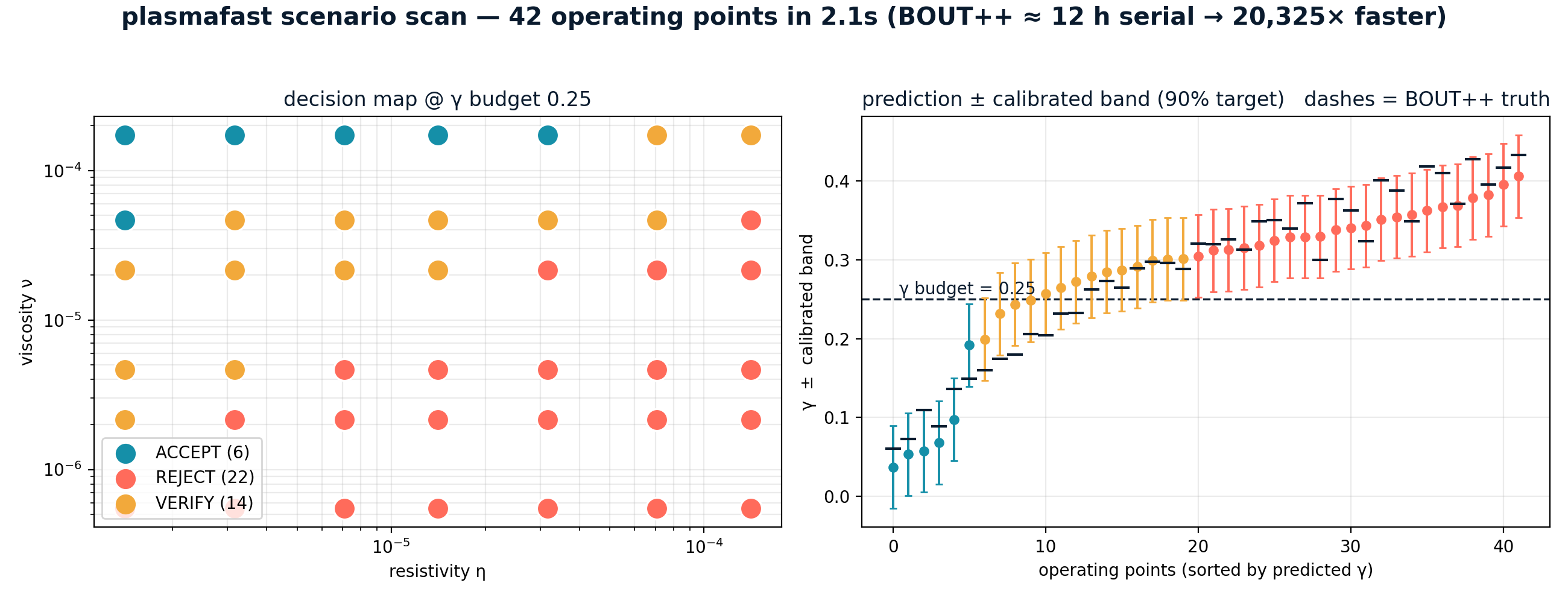}
\caption{Scenario scan on 42 held-out BOUT++ operating points at a growth-rate budget of 0.25.
Left: decision map in $(\eta,\nu)$. Right: prediction $\pm$ calibrated interval, sorted by
prediction, with BOUT++ values as dashes.}
\label{fig:scan}
\end{figure*}

\paragraph*{Scenario scan.} Figure~\ref{fig:scan} screens the 42 held-out operating points
against a growth-rate budget: a point is accepted if its whole interval is below the budget,
rejected if above, and sent to BOUT++ otherwise (intervals fitted leave-one-out). The scan takes
about 2\,s; BOUT++ took a median 16.7\,min per point on x86 CPUs, about 12\,h serially. Across
budgets from 0.05 to 0.45, 289 confident decisions contain no error; 43--98\% of the scan is
settled without BOUT++, least when the budget sits inside the dense high-$\gamma$ band.

\section{Lessons and corrections}\label{sec:lessons}

We report the problems found while preparing this work, because each would have misled a user.
\begin{itemize}
\item \emph{An irreproducible number.} An earlier version of this work (our conference
abstract) reported that ensemble spread rank-correlates with error at 0.69. The ensemble had
been scored before one member finished training; with all members trained the correlation is
about zero in distribution. In-distribution errors are nearly uniform, so a fixed conformal
interval is already the right tool, and we use the spread only as an OOD alarm.
\item \emph{Inputs the network could not see.} Resistivity and viscosity entered as raw
constant channels of order $10^{-6}$--$10^{-4}$ next to order-one channels; the growth-rate head
saturated at 0.33 for every case above 0.25, on training data too. Feeding $\log_{10}\eta$ and
$\log_{10}\nu$ to the head halved the held-out error (0.053 to 0.028) and the interval width.
\item \emph{A validation gate on the wrong case.} A check against PB3D read the input of a
different run because of an indexing error; fixed, it also showed that comparing a resistive
surrogate with an ideal code is the wrong test, so we judge the surrogate against BOUT++ and
report BOUT++ against PB3D separately.
\item \emph{Missing global inputs.} The deployed input keeps only profile shapes; plasma shape
and pressure magnitude are not seen. This is a main reason the BOUT++ surrogate fails on unseen
machines, and the motivation for the v2 model.
\item \emph{A grid too coarse at the edge.} The deployed input samples $q$ and $p$ on 64
points uniform in $\rho_{\rm pol}$, an edge spacing of $\Delta\psi_N = 0.031$; the steepest
pressure gradient in KARHU sits at $\psi_N \approx 0.975$--$0.99$, so the pedestal gets one or
two points. The resampling alone costs a simple MLP 16\% in mean error (three seeds). An
edge-packed grid (32 points on $\psi_N\in[0,0.8)$, 64 on $[0.8,1]$) recovers it
(Sec.~\ref{sec:karhu}).
\item \emph{The same scaling mistake, twice.} After the $\eta$, $\nu$ lesson above, the $q$ and
$p$ profiles still entered raw. Standardizing them per radial point cut a single network's
error by 18\%, more than the grid fix; the two together made the KARHU surrogate competitive.
\item \emph{Two input conventions.} The BOUT++ training frames, built from simulation output,
and the frames built from a G-EQDSK at inference differ: the radial coordinate spans the
simulation annulus ($\psi_N$ 0.40--1.19) in one and axis to separatrix in the other, the
safety-factor channel is empty in training, and pressure and poloidal angle are normalized
differently. Real-machine predictions therefore rest on the four fine-tuning frames, which
is consistent with the large error on JET. The held-out and scenario-scan results use the
training convention and are unaffected. One frame builder for training and inference is the
fix; the KARHU surrogates already use one, and the shared builder regenerates their training
frames byte for byte.
\item \emph{Which growth rates to trust.} Our first quality rule for BOUT++ labels rejected 46 of
64 SPARC runs whose toroidal harmonics grew at one rate. Read on the properly assembled grid, the
runs seed only $n = 15$; the other harmonics start at round-off ($10^{-16}$ of it) and copy its growth,
so the label is the seeded mode's rate and the copies are ignored. The rates of 1--3 that worried
us came from a numerical mode at the inner radial boundary, mostly at the lowest resistivity; such
runs are now excluded rather than labelled stable. The trainable SPARC set grew from 14 to 51
labels from the same runs, and an equilibrium's growth rate agrees across resistivities to a
median factor of 1.2.
\item \emph{Dividing by a near-zero spread.} Standardizing each radial point by the training
spread failed on real equilibria: near the axis KARHU's pressure profiles are almost identical,
so a small real difference became $|z| \approx 100$ and the ensemble members diverged. A floor on
the spread (10\% of the channel mean) removed that without changing accuracy on KARHU.
\end{itemize}

\section{Discussion and limitations}

For a group designing a new device, the KARHU results translate into two ways to start trusting
a surrogate on their equilibria: about thirty simulations of the new design to recalibrate, or
none, at the price of sending roughly one case in four to the physics code. Every routed case
also produces a label where the surrogate is weakest, so routing and recalibration combine into
an active-learning loop~\cite{zanisi2024active}.

The guarantees are marginal and assume exchangeability between calibration and use; no
distribution-free method can certify an individual prediction, which is why routing to the
physics code remains part of the design. The KARHU test covers one machine family, one shift
variable and ideal MHD. The BOUT++ surrogate is trained on essentially one equilibrium plus
four real cases, so it is in effect a model of $\gamma(\eta,\nu)$ on CBM18, and its
real-machine alarm result rests on six equilibria; cross-machine skill needs many
equilibria, as in the KARHU surrogates. The input fixes found on KARHU (edge-packed grid,
standardized profiles, one frame builder) are not yet applied to the BOUT++ surrogate; the v4 in-distribution result is selection-biased, and only its shifted-split result is confirmatory. On real machines we have no MISHKA labels, so
the real-equilibrium result is about refusing correctly, not about accuracy. Nothing here is
validated against experiment; anchoring to observed ELM onset is the next step. The approach
is model-agnostic and could wrap other surrogates, ELMO~\cite{li2026elmo} and
KARHU~\cite{bruncrona2025karhu} included.

\begin{acknowledgments}
We thank A.~Bruncrona, A.~Kit and A.~J\"arvinen (VTT) for making the KARHU training data public
under CC BY 4.0. This work was supported by dataquanta LLC.
\end{acknowledgments}

\section*{Data and code availability}
KARHU data: Zenodo, doi:10.5281/zenodo.18629219. Analysis code, pre-registration and results
for the KARHU experiments are available from the author on request (amin@dataquanta.io).

\bibliography{refs}

\end{document}